%% file: neurips_2024.tex
\documentclass{article}

\usepackage[preprint]{neurips_2024}

\usepackage[utf8]{inputenc} 
\usepackage[T1]{fontenc}    
\usepackage{hyperref}       
\usepackage{url}            
\usepackage{booktabs}       
\usepackage{amsfonts}       
\usepackage{nicefrac}       
\usepackage{microtype}      
\usepackage{xcolor}         

\usepackage{graphicx}
\usepackage{amsmath}
\usepackage{amssymb}
\usepackage{multirow}
\usepackage{makecell}
\usepackage{diagbox}
\usepackage{pifont}
\usepackage{colortbl}
\usepackage{todonotes}
\usepackage{subcaption}
\usepackage{array}

\newlength\savewidth
\newcommand{\tablestyle}[2]{\setlength{\tabcolsep}{#1}\renewcommand{\arraystretch}{#2}\centering\footnotesize}

\newcommand{\cmark}{\ding{51}}
\newcommand{\xmark}{\ding{55}}

\usepackage{soul}

\definecolor{deemph}{gray}{0.6}

\definecolor{backgroundcolor}{RGB}{232, 242, 255}
\definecolor{mygray}{gray}{0.9}
\newcommand{\mygrayhl}[1]{\sethlcolor{mygray}\hl{#1}}

\newcommand{\method}{ILLUME+}
\newcommand{\tokenizer}{DualViTok}

\title{\includegraphics[width=0.05\textwidth]{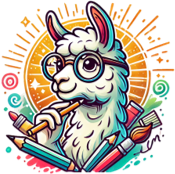} \method: Illuminating Unified MLLM with Dual Visual Tokenization and Diffusion Refinement}

\author{
    Runhui Huang$^{2}$$^{\ast}$\quad Chunwei Wang$^{1}$$^{\ast}$\quad Junwei Yang$^{1}$\quad Guansong Lu$^{1}$\quad \\ \textbf{Yunlong Yuan}$^{1}$ 
     \quad \textbf{Jianhua Han}$^{1}$\quad \textbf{Lu Hou}$^{1}$ \quad \textbf{Wei Zhang}$^{1}$ \\ \textbf{Lanqing Hong}$^{1}$ \quad \textbf{Hengshuang Zhao}$^{2}$$^{\dag}$ \quad \textbf{Hang Xu}$^{1}$$^{\ddag}$    \\
    \textit{$^{1}$Huawei Noah’s Ark Lab, $^{2}$The University of Hong Kong}
}

\begin{document}
\maketitle
\footnotetext{$^{\ast}$Equal contribution, $^{\dag}$Corresponding author, $^{\ddag}$ Project leader.} 

\begin{figure*}[h]
\vspace{-1cm}
\centering
\includegraphics[width=0.95\textwidth]{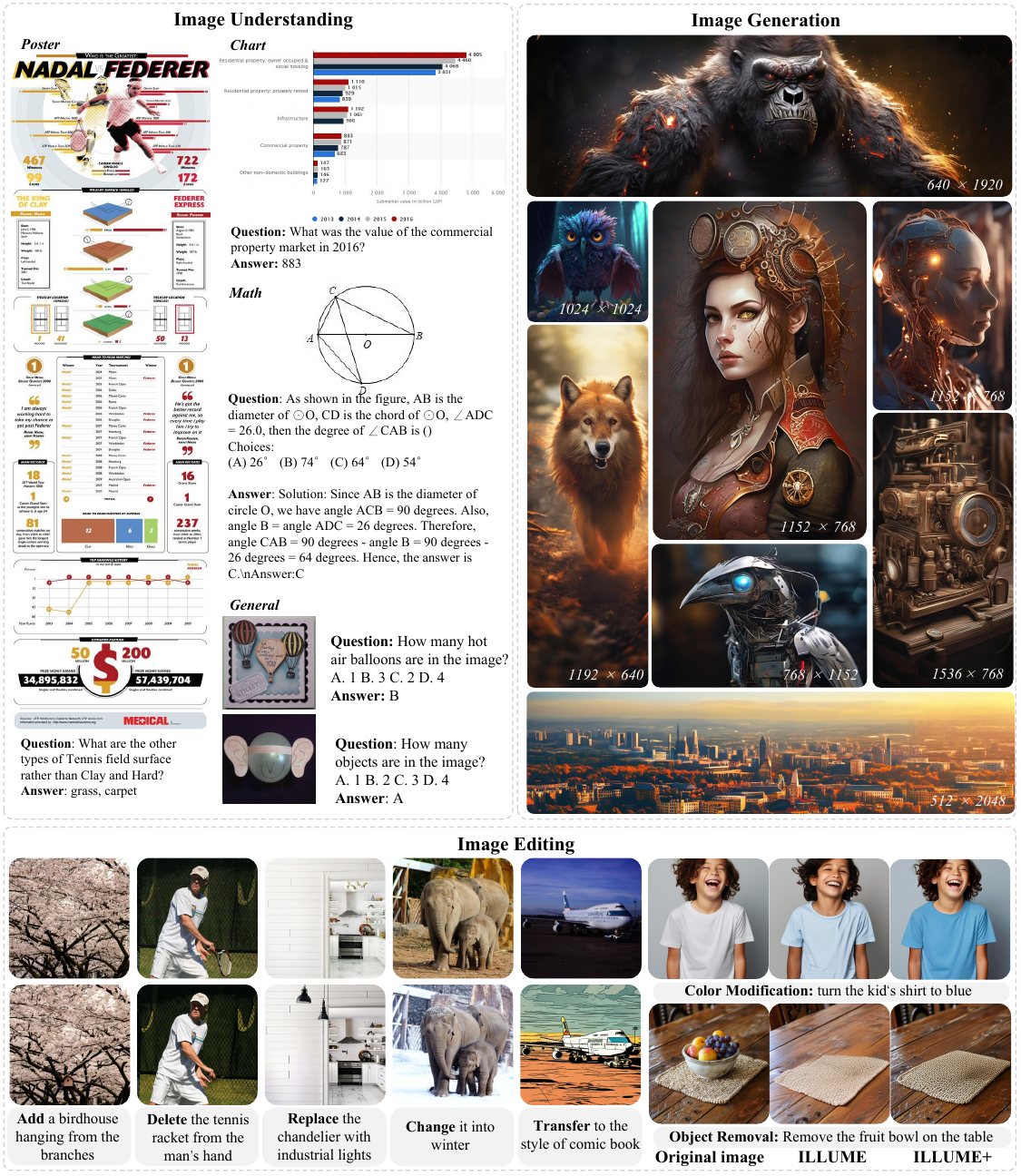}
\caption{\method~can understand and generate images at any resolution. Compared to our previous work, ILLUME~\cite{wang2024illume}, it demonstrates improved texture preservation in image editing tasks.}
\vspace{-3mm}
\label{fig:vis}
\end{figure*}

\newpage
\input{sec/0_abstract}    
\input{sec/1_intro}

\input{sec/2_related_work}

\input{sec/3_method}

\input{sec/4_experiments}
\input{sec/5_conclusion}

\clearpage
{
\small
\bibliographystyle{abbrv}
\bibliography{neurips_2024}
}




\end{document}

%% file: sec/0_abstract.tex
\begin{abstract}
We present \textbf{\method}, an enhanced version of the previous ILLUME model, which leverages dual visual tokenization and a diffusion decoder to improve both deep semantic understanding and high-fidelity image generation.
Existing unified models have struggled to simultaneously handle the three fundamental capabilities expected of a unified model: understanding, generation, and editing. 
Models like Chameleon and EMU3 utilize VQGAN for image discretization. Due to the lack of deep semantic interaction, they lag behind specialist models like LLaVA in visual understanding tasks.
LaViT and ILLUME employ semantic encoders for tokenization, but they struggle with image editing due to poor texture preservation. 
Meanwhile, Janus series decouples the input and output image representation, limiting their abilities to seamlessly handle interleaved image-text understanding and generation.
In contrast, \method~introduces a unified dual visual tokenizer, \tokenizer, which preserves both fine-grained textures and text-aligned semantics while enabling a coarse-to-fine image representation strategy for multimodal understanding and generation. Additionally, we employ a diffusion model as the image detokenizer to enhance generation quality and enable efficient super-resolution. \method~follows a continuous-input, discrete-output scheme within the unified Multimodal Large Language Model (MLLM) and adopts a progressive training procedure that supports dynamic resolution across the vision tokenizer, MLLM, and diffusion decoder. This design allows for flexible and efficient context-aware image editing and generation across diverse tasks.
\method~(3B) exhibits competitive performance against existing unified MLLMs and specialized models across multimodal understanding  generation, and editing benchmarks. Its support for flexible high-resolution images enhances visual understanding tasks and enables detailed image synthesis up to 1024×1024 resolution. 
With its strong performance, \method~provides a scalable foundation for future multimodal model applications.
Project Page: \url{https://illume-unified-mllm.github.io/}. Code and models will be publicly available soon.

\end{abstract}

%% file: sec/1_intro.tex
\section{Introduction}
\label{sec:intro}

{\centering “What I cannot create, I do not understand.”
——Richard Feynman \par}

Recent advancements in Large Language Models (LLMs) have significantly enhanced their capability to handle multimodal tasks, particularly by integrating visual inputs into language models. 
Efforts such as the LLaVA series and the QwenVL series~\cite{liu2024llava,liu2024improved,bai2023qwen,wang2024qwen2}  have demonstrated remarkable visual comprehension performance.
Meanwhile, the development of text-to-image generation models, such as diffusion-based approaches~\cite{rombach2022high,peebles2023scalable,podell2023sdxl,ruiz2023dreambooth} and more recent autoregressive approaches
~\cite{ramesh2021zero,ding2021cogview,yu2022scaling}, has made substantial strides in generating high-fidelity images. 
These developments have driven the push towards creating unified Multimodal Large Language Models (MLLMs) that seamlessly integrate both visual understanding and generation capabilities.
A unified model holds promise not only for advancing task coordination and generalization but also for contributing to the exploration of artificial general intelligence (AGI). 
By merging understanding and generation capabilities within a single framework, unified models can genuinely grasp the deep relationships between visual and textual information, enabling more intelligent and flexible interactions and task execution in complex real-world scenarios.

\begin{figure*}[t]
\centering
\includegraphics[width=\textwidth]{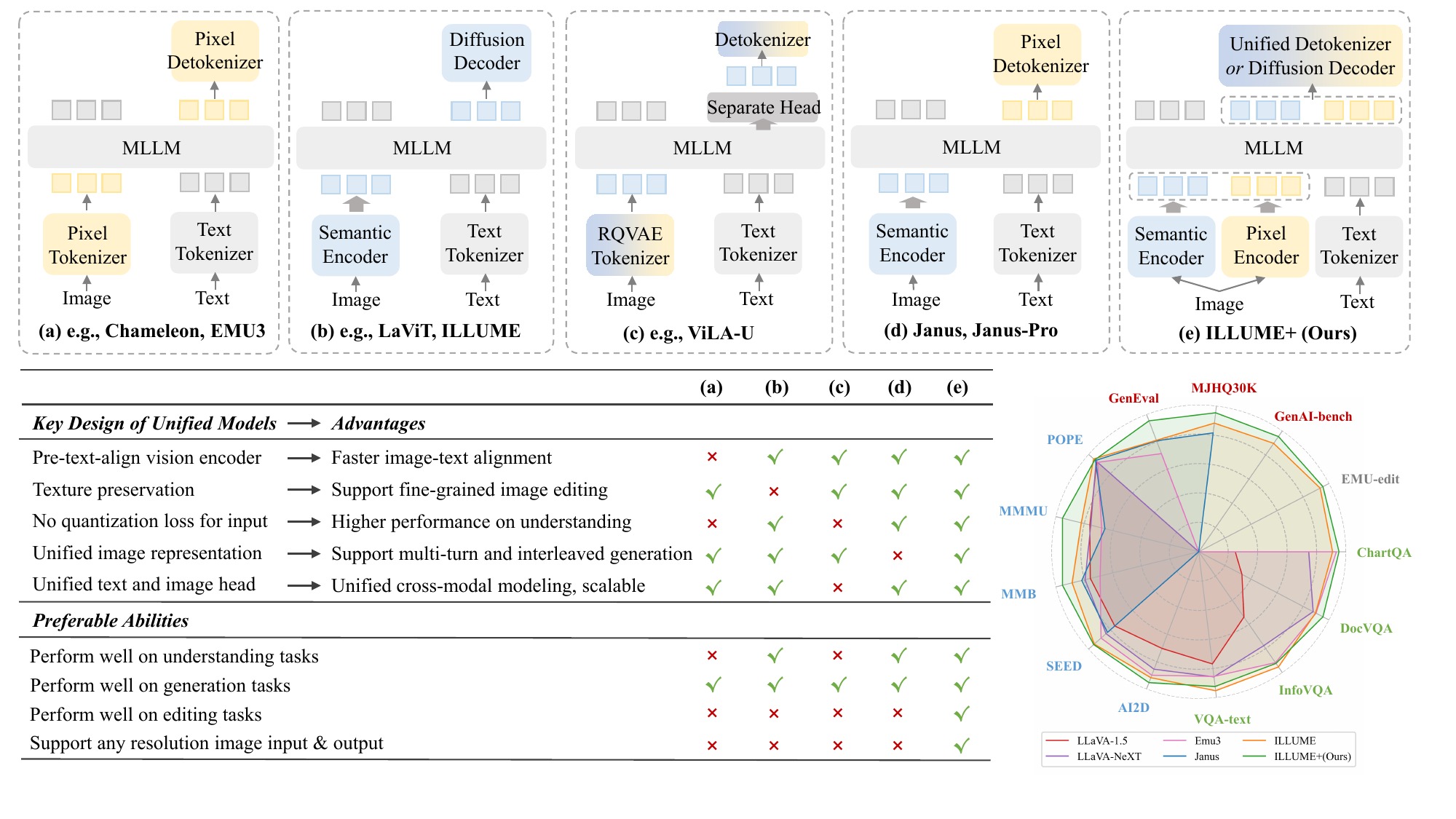}
\caption{Characteristics comparison among existing unified models. Existing methods explore distinct paradigms to balance visual understanding, generation, and editing capabilities. Early approaches using VQGAN discretization struggle in understanding and context-aware generation tasks due to limited semantic alignment. Later frameworks incorporate semantic encoders, achieving better alignment but compromising texture preservation essential for fine-grained editing. ILLUME+ deep-integrates image understanding, generation, and editing into a single, unified architecture, enabling more intelligent and flexible interactions and task execution.}
\vspace{-3mm}
\label{fig:intro}
\end{figure*}

To build such unified models in an autoregressive framework, existing approaches have explored several distinct paradigms.
As illustrated in Fig.~\ref{fig:intro} (a), the earliest models, e.g., Chameleon~\cite{team2024chameleon} and Emu3~\cite{wang2024emu3}, utilized VQGAN~\cite{esser2021taming} to discretize images, enabling a shared vocabulary for text and vision within an autoregressive (AR) framework. However, even with large-scale training, they lag behind models like LLaVA~\cite{liu2024llava} in visual understanding tasks.
To address this issue, works like LaViT~\cite{jin2024unified} and ILLUME~\cite{wang2024illume} (Fig.~\ref{fig:intro} (b)) learn the semantic codebook from the pretrained text-aligned semantic encoders~\cite{zhai2023sigmoid,zhu2024unit}, and employ diffusion models to reconstruct images based on semantic tokens. This improves image-text alignment in MLLM pretraining and achieves strong performance in both understanding and generation tasks. However, the lack of texture preservation in vision tokenizers limits their capability in fine-grained image editing.
To mitigate this, another line of work (Fig.~\ref{fig:intro} (c)) integrates pre-aligned vision encoders with image-text contrastive learning during VQGAN training. 
Due to the usage of RQVAE to balance the pixel reconstruction and image-text alignment, it requires a separate head for image generation, increasing design complexity and potentially posing a bottleneck when scaling up the models. 
Finally, methods in Fig.~\ref{fig:intro} (d) decouple image understanding and generation tasks by employing the semantic encoder and VQGAN tokenizer independently. While effective, this structure fails to support interleaved image-text tasks and multi-turn dialogues, as it necessitates manual specification of whether a task involves understanding or generation, contradicting the flexibility expected of a unified foundation model.

By reviewing the current research, we suggest that \textit{\textbf{a strong unified foundation model should demonstrate three core competencies}}: \textit{\textbf{visual understanding}} (accurate image interpretation), \textit{\textbf{generation}} (high-quality image synthesis), and \textit{\textbf{editing}} (instruction-following modification while maintain consistency of other parts). These are fundamental capabilities that guarantee the model has the potential to benefit from scaling in both model capacity and task diversity. Then it raises a question: \textit{\textbf{How can we construct such a unified foundation model?}} We summarize the following key design principles:

\underline{\textit{Pre-text-align vision encoder.}} Previous studies~\cite{lu2024deepseekvl,tong2024cambrian} have shown that the pre-text-aligned vision encoder, i.e., CLIP-like models, significantly benefit to visual understanding capability. Additionally, in contrast to VQGAN-based models (Fig.~\ref{fig:intro} (a)) that purely supervised by image reconstruction, these semantic encoders facilitate image-text alignment as demonstrated in ILLUME~\cite{wang2024illume}. Thus, incorporating a pre-text-aligned semantic encoder is essential for a unified foundation model.

\underline{\textit{Image texture preservation.}} The quality of image reconstruction of vision tokenizers is essential for handling editing tasks, which determines the up-bound of unified models to maintain consistency of unchanged regions in images. Therefore, not only semantic information but also texture preservation is required as a crucial consideration in vision tokenizer design choices.

\underline{\textit{No information loss for image input.}} While the vision tokenizer is the key to enabling unified autoregressive image-text generation, it inevitably introduces information loss during the quantization process. To this end, using continuous features before the quantizer of vision tokenizer as visual input for LLM serves as a more suitable choice to guarantee fine-grained multimodal understanding capability. 

\underline{\textit{Unified image input and output representation.}} The decoupled mechanism in Janus series~\cite{wu2024janus,chen2025janus_pro}, i.e., semantic representation for visual input while pixel representation for visual output, inevitably hinders the model’s ability to accurately interpret and further modify its own visual outputs in multi-round steps. Hence, a unified representation for visual input and output is necessary to further support image-text interleaved generation, multi-turn dialogues and chain-of-thought reasoning.

\underline{\textit{Unified text and image head.}} In a single autoregressive framework, a unified output head for both image and text is preferable, as it not only simplifies infrastructure design but also enhances cross-modal interactions. In contrast, the requirement of a separate image head in Fig.~\ref{fig:intro} (c) introduces challenges in modality switching during generation. For example, as we need special line separator tokens interleaved with visual tokens to represent an image in different resolutions, how can the model seamlessly transition between text and image heads during inference? To avoid such complexity, a unified head offers a more effective and elegant solution.

Building on the above analysis, we introduce \method, an enhanced version of the ILLUME model that encompasses all the key designs mentioned above but also exhibits all the preferable abilities, listed in Fig.~\ref{fig:intro}'s Table). \method~supports flexible any-resolution visual input and output and excels in multimodal understanding, generation, and editing tasks, as demonstrated in Fig.~\ref{fig:vis}. Its key features are outlined below:

\textbf{Dual vision tokenizer for semantic and texture preservation.} We introduce the \tokenizer, a dual-branch vision tokenizer designed to capture both deep semantics and fine-grained textures. The semantic branch utilizes a pre-trained text-aligned vision encoder for semantic feature extraction, supervised by feature reconstruction loss. In parallel, the pixel branch integrates quantized features from both the semantic encoder and a CNN-based pixel encoder to enhance pixel-level reconstruction. To improve robustness against incorrect token predictions in autoregressive generation, we introduce noise injection during training by randomly perturbing visual tokens. Despite its simplicity, \tokenizer~is specifically designed for unified models, ensuring both semantic and texture preservation while maintaining robust token decoding.

\textbf{Unified MLLM with unified coarse-to-fine image representation.} Unlike the Janus series~\cite{wu2024janus,chen2025janus_pro}, which decouples visual input and output representations, we adopt a coarse-to-fine strategy, first generating semantic tokens followed by pixel tokens. This sequential arrangement enables LLMs to utilize a unified LM head with a simple vocabulary expansion while leveraging semantic visual tokens as a bridge to enhance alignment between text and visual textures. Additionally, to prevent information loss at the input stage, we employ a continuous-input, discrete-output scheme following ILLUME~\cite{wang2024illume}, using pre-quantized continuous features as inputs while generating discrete tokens for image synthesis. 

\textbf{Diffusion decoder for enhanced generation quality and efficient super-resolution.} We incorporate a diffusion model as an optional choice for image generation, offering two key benefits: (i) Higher generation quality. Diffusion models refine details and reduce artifacts, surpassing direct token decoding from a vision tokenizer in both fidelity and robustness. (ii) Efficient super-resolution. They upscale images during decoding, mitigating the token explosion issue in autoregressive high-resolution generation. 

\textbf{Progressive training procedure for flexible resolution visual input and output.} We employ a progressive training procedure for all the above three modules, gradually increasing resolution from fixed low to flexible high, to ensure training stability and final performance. Additionally, during MLLM training, we incrementally increase tasks diversity and complexity, with carefully designed data distribution for each stage.

Based on these design choices, \method~with only a 3B LLM excels among existing unified MLLMs and exhibits competitive performance against specialized models across multimodal understanding, generation, and editing benchmarks. Its support for high-resolution images enhances both visual understanding and detailed image synthesis, surpassing existing unified models in document-oriented tasks and enabling generation up to 1024×1024 resolution. 
Additionally, \method~improves texture preservation over ILLUME in image editing as illustrated in Fig.~\ref{fig:vis}. 
\method's robust performance across these diverse benchmarks highlights its effectiveness in unifying understanding, generation, and editing, providing a scalable solution for future multimodal applications.

%% file: sec/2_related_work.tex
\section{Related Work}
\label{sec:related_work}

\noindent \textbf{MLLM for image understanding.} Recent advancements in Large Language Models (LLMs) have led to the development of Multimodal Large Language Models (MLLMs) for image understanding. Early models like LLaVA \cite{liu2024visual} and MiniGPT-4 \cite{zhu2023minigpt} used vision adapters to align visual features with LLMs, showing strong performance in visual perception tasks. Later models such as Qwen-VL series~\cite{bai2023qwen,wang2024qwen2}, and InternVL series~\cite{chen2024internvl,chen2024far} improved upon this with higher-quality datasets and better training strategies, but still focus primarily on visual understanding. 
However, despite the strong understanding capabilities of these models, they primarily focus on visual perception and comprehension. This highlights the need for more comprehensive solutions that unify both understanding and generation, allowing models to learn deeper relationships between visual and textual information, enabling more intelligent and flexible interactions and task execution in complex real-world scenarios.

\noindent \textbf{Image generation model.}
Generative adversarial networks (GANs \cite{goodfellow2014gan}) is the pioneering method for image generation in the era of deep learning. 
However, it suffers from the problem of unstable training process and mode collapse. In recent years, diffusion-based methods~\cite{ddpm,rombach2022high,peebles2023scalable,ruiz2023dreambooth,betker2023improving} have shown excellent image generation capabilities. These models learn to predict Gaussian noise in a forward diffusion process, and then generate high-quality images through an inverse denoising process. 
Among them, the latent diffusion model \cite{rombach2022high,podell2023sdxl,stable_diffusion_3} addresses computational challenges by creating images from low-dimensional latent representations.
Another line of research turns to explore the image generation in the autoregressive model~\cite{ramesh2021zero,ding2021cogview,yu2022scaling,var}, which converts images into discrete tokens using VQGAN-like vision tokenizers~\cite{esser2021taming,lee2022autoregressive} and generates images by predicting the tokens autoregressively. In this paper, we unify the image generation and image editing into one MLLM with image understanding tasks under the autoregressive manner, and further adopt a diffusion model to further improve the reconstruct quality from the predicted tokens.

\noindent \textbf{Unified multimodal understanding and generation.} Early efforts to unify visual understanding and generation using LLMs include models like Emu~\cite{sun2023generative} and X-VILA~\cite{ye2024x}, which adopt unified autoregressive approaches to predict multimodal elements. However, the non-unified optimization of different modalities limits feature integration, and additional components like diffusion decoders reduce efficiency. Models such as LWM~\cite{liu2024world}, Chameleon~\cite{team2024chameleon}, and VILA-U~\cite{wu2024vila} use VQ tokenizers to convert images into vision tokens, enabling a unified training framework for text and image generation. Despite these advancements, challenges remain in integrating understanding and generation. Janus series~\cite{wu2024janus,chen2025janus_pro} decouples visual encoding for understanding
and generation, which may suffer from misaligned representations due to separate branches for understanding and generation. These limitations highlight the need for better solutions that allowing flexible and efficient context-aware image understanding
and generation across various tasks.

%% file: sec/3_method.tex
\section{Method}
\label{sec:method}
\begin{figure*}[t]
\centering
\includegraphics[width=\textwidth]{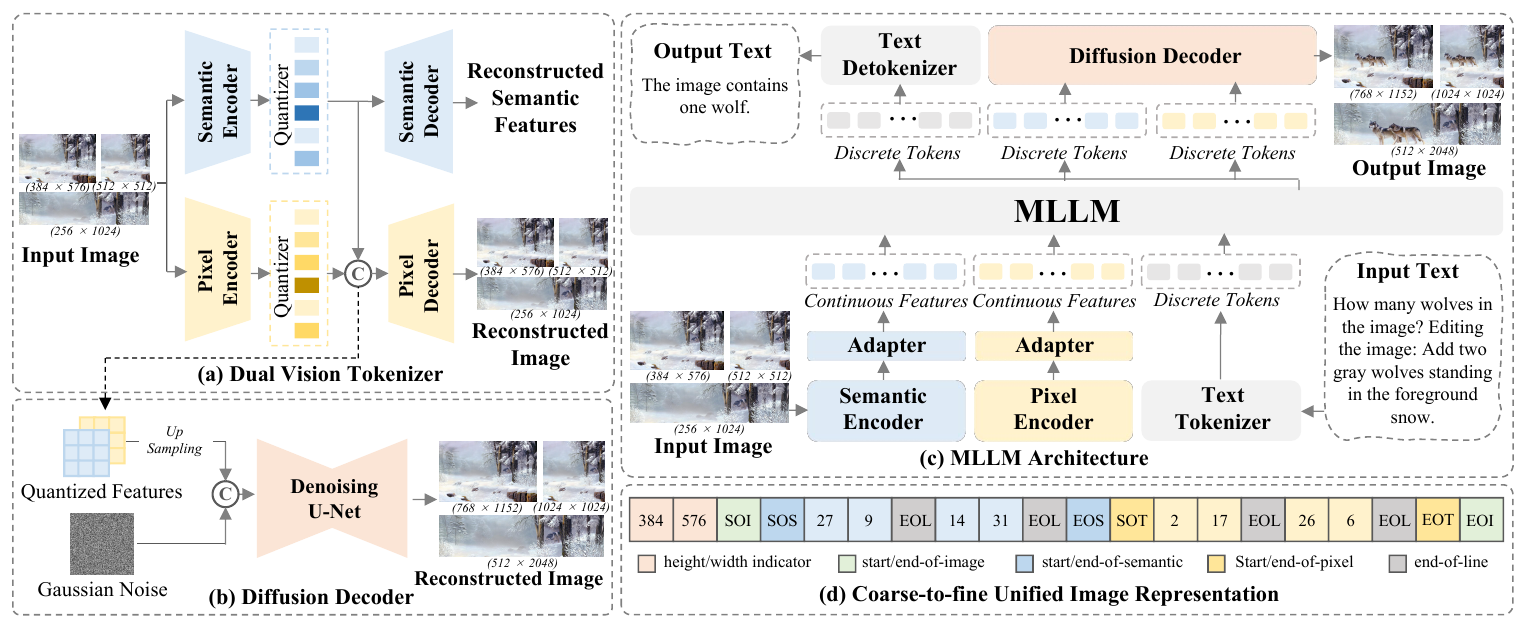}
\caption{\textbf{Architecture of \method.}
(a) The dual vision tokenizer preserves both semantic and texture information. 
(b) The diffusion refiner decodes discrete tokens into high-quality images. 
(c) The unified MLLM enables deep semantic interactions and context-aware image generation. 
(d) We introduce an unambiguous image representation of discrete tokens in a chain-of-thought pattern (semantic tokens first, followed by pixel tokens), resulting in improved generation performance.} 
\vspace{-3mm}
\label{fig:architecture}
\end{figure*}

Figure~\ref{fig:architecture} provides an overview of our proposed framework, \method, which comprises a dual vision tokenizer, a MLLM, and a diffusion decoder. 
Our architecture's core design principle is the unified dual visual tokenization mechanism that captures both deep semantic information and fine-grained texture details, ensuring a comprehensive image representation for visual understanding, generation, and editing tasks.
The following section elaborates on the architectural details, training procedures, and data composition.

\subsection{Dual Vision Tokenizer}
\label{sec:dual_vision_tokenizer}
Although vision tokenizers have been studied for years, their design for unified models that simultaneously support understanding, generation, and editing remains an open challenge. 
To mitigate this, we propose the Dual Vision Tokenizer~(\tokenizer) tailored for unified models with semantic and texture preservation and robust decoding.

\textbf{Semantic and texture information preservation.} As shown in the Fig.~\ref{fig:architecture} (a), \tokenizer~incorporates a dual-branch design to learn both deep semantics and fine-grained textures. 
The semantic branch leverages the pre-trained text-aligned vision encoder, QwenViT~\cite{wang2024qwen2}, to extract high-level semantic features, which are then quantized into discrete tokens and reconstructed using a lightweight decoder. 
The semantic reconstruction is optimized with a cosine similarity loss between the reconstructed features and semantic features from the vision encoder.
Meanwhile, the pixel branch follows a MoVQGAN-based architecture~\cite{zheng2022movq}. 
After quantization, the pixel and semantic tokens are concatenated along the channel dimension for image decoding. Following standard VQGAN procedures~\cite{esser2021taming}, the pixel branch is trained with L1 loss, perceptual loss, and GAN loss.
More specifically, for the semantic branch, we adopt a 28× downsampling rate commonly used in state-of-the-art understanding models~\cite{li2024llavaonevision,wang2024qwen2} and employ a 16× rate for the pixel branch to preserve fine-grained textures. 
To further minimize the information loss induced by tokenization, we incorporate space-to-channel and channel-to-space transformations inspired by DC-AC~\cite{chen2024dcae} in the downsampling and upsampling process. 
Additionally, to ensure reconstruction fidelity, we employ larger codebook sizes compared with previous works~\cite{esser2021taming}, 32,768 for the semantic branch and 98,304 for the pixel branch, with SimVQ~\cite{zhu2024simvq} as quantization method to maintain high codebook utilization rate. Please refer to ablation studies in Sec.~\ref{sec:experiments} for more discussions about detailed design choices of our vision tokenizer.

\textbf{Robust decoding for incorrect token predictions.} 
LLMs may occasionally predict incorrect visual tokens, and the tokenizer decoder should be resilient to such errors to minimize artifacts in the generated images. 
To improve robustness, we introduce noise injection during tokenizer training: each sample has an $\alpha$=10\% probability of being perturbed, with $\beta$=10\% of its tokens randomly replaced, helping the decoder better handle erroneous token predictions.

\subsection{Unified Multimodal Large Language Model}
As shown in Fig.~\ref{fig:architecture} (c), \method~inherits the architecture of existing VLMs~\cite{liu2024improved,liu2024llava} by extending LLMs with an additional vision vocabulary to generate discrete vision tokens, employing a continuous-input, discrete-output scheme for  image processing. To avoid information loss caused by tokenizer quantization, we employ both the semantic encoder and pixel encoder within dual vision tokenizers to extract features from input images, which are then aligned with the LLM’s input space via two separate vision adaptors. 
For visual generation, images are converted into discrete tokens, enabling a unified modeling of visual and text tokens within the LLM. 
By using the same next-token prediction loss, both modalities are seamlessly integrated into a shared prediction head.

During inference, we apply the classifier-free guidance (CFG) following~\cite{xie2024show, wang2024illume} for text-to-image generation and image editing tasks, where the unconditioned setting serves as masked text descriptions and editing instructions, respectively.

\textbf{Coarse-to-fine unified image representation.} 
As illustrated in Fig.~\ref{fig:architecture} (d), we adopt a unified representation for images with a coarse-to-fine sequence arrangement, i.e., first generating semantic tokens followed by pixel tokens. Since semantic representations align more naturally with text, generating semantic tokens first allows the model to determine content before refining details based on semantic information, thus enhancing alignment between text and visual textures.
Specifically, we use <start-of-image/semantic/pixel> and <end-of-image/semantic/pixel> markers to indicate the boundaries of the entire image, semantic representation, and texture representation, respectively. 
Additionally, <end-of-line> tokens are inserted at the end of each row to distinguish different resolutions, while the height and width indicators at the sequence's start provide explicit resolution information during image generation.

\subsection{Diffusion Decoder}
We introduce an additional diffusion model to decode image from predicted discrete tokens for enhanced generation quality and efficient super-resolution. 
Specifically, based on the SDXL model~\cite{podell2023sdxl}, our diffusion decoder replaces text encoders with zero embeddings in the cross-attention layers. Semantic and pixel tokens from dual vision tokenizer are mapped to feature representations using the learned codebooks and then injected into the UNet model by concatenation with the noisy image latent. 
Note that the model performs super-resolution, doubling the image size (i.e., 256×256 to 512×512), to mitigate the token explosion issue in autoregressive high resolution generation.
During training, similar to our dual vision tokenizer, random perturbations are applied to 50\% of the samples, replacing 10\% of the tokens to enhance robustness against noisy input tokens. 
To prioritize semantic tokens, we use a smaller masking probability (10\%) for semantic features and larger (50\%) for texture.

\begin{figure*}[t]
\centering
\includegraphics[width=\textwidth]{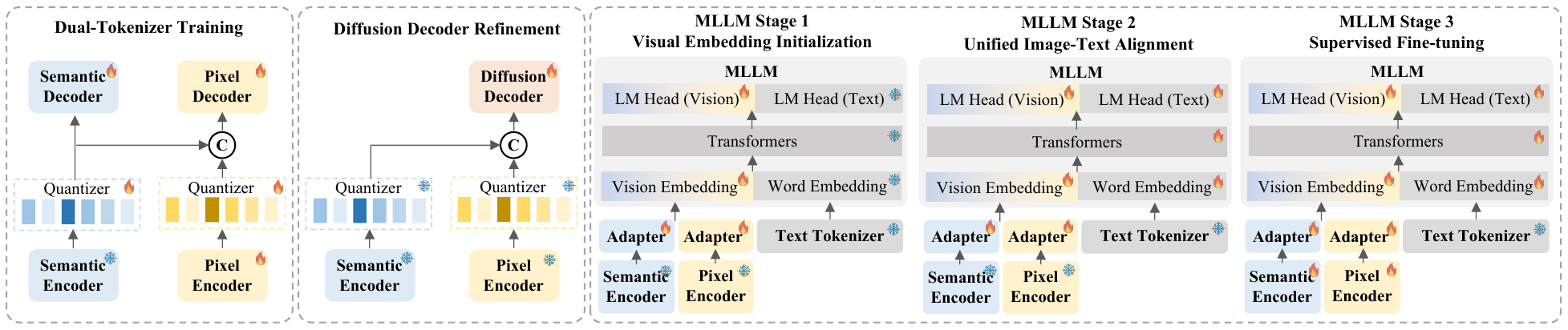}
\caption{Illustration of our progressive training pipeline. We first pre-train the dual-tokenizer system by reconstruction of the semantic and pixel information. We then fine-tune the diffusion model as a high-quality image decoder. The MLLM training consists of three main stages that gradually increase task resolution and complexity.}
\vspace{-3mm}
\label{fig:training_procedure}
\end{figure*}

\subsection{Training Procedure and Data Composition}
\begin{figure}[t]
    \centering
    \begin{minipage}{0.56\textwidth} 
        \centering
        \includegraphics[width=\textwidth]{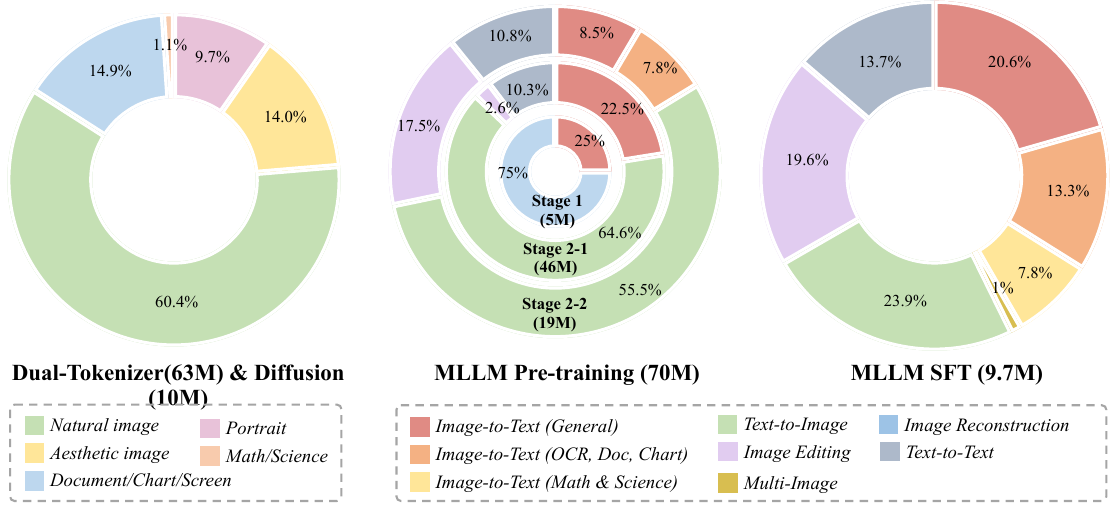}
    \end{minipage}
    \begin{minipage}{0.42\textwidth} 
        \centering
        \tiny
        \resizebox{1.0\textwidth}{!}{
        \begin{tabular}{@{} l p{4.2cm} @{}}
            \toprule
            \textbf{Stage} & \textbf{Dataset} \\
            \midrule
            Tokenizer \& Diffusion & COYO~\cite{kakaobrain2022coyo-700m}, EMOVA, in-house aesthetics data  \\
            \midrule
            \multirow{5}{*}{MLLM Pre-training} & COYO, Wukong~\cite{gu2022wukong}, EMOVA-Pretrain~\cite{chen2024emova}, \\
            & LLAVA-SFT~\cite{liu2024visual}, in-house aesthetics data, \\
            & UltraEdit~\cite{zhao2024ultraedit}, SEED-Edit~\cite{ge2024seed}, AnyEdit~\cite{yu2024anyedit}  \\
            & Magpie~\cite{xu2024magpie}, OpenOrca~\cite{OpenOrca}, SCP-116K~\cite{lu2025scp}, \\
            & OpenHermes~\cite{OpenHermes2.5}, OPC-SFT-Stage1~\cite{Huang2024OpenCoderTO} \\
            \midrule
            \multirow{5}{*}{MLLM SFT} & EMOVA-SFT, Pixmo~\cite{deitke2024molmo}, M4-Instruct~\cite{liu2024improved}  \\ 
            & COYO, in-house aesthetics data, \\
            & OmniEdit~\cite{wei2024omniedit}, AnyEdit~\cite{yu2024anyedit}, UltraEdit~\cite{zhao2024ultraedit}, \\
            & Instruct-Pix2Pix~\cite{brooks2023instructpix2pix},\\
            & Magpie~\cite{xu2024magpie}\\
            \bottomrule
        \end{tabular}}
        \end{minipage}
    \caption{Summary of the data mixture in each stage. Our training data gradually covers a wide range of tasks and various image resoluton.}
    \label{fig:data_composition}
\end{figure}

Figure~\ref{fig:training_procedure} illustrates the whole training procedure of our proposed \method, where we adopt a progressive training paradigm designed to support flexible resolutions and leverage the capabilities of each component. An overview of data distribution at each stage is illustrated in Fig.~\ref{fig:data_composition}. 

\noindent \textbf{Dual vision tokenizer training.} As illustrated in Fig.~\ref{fig:training_procedure}, all components of our proposed \tokenizer~except for the pretrained semantic encoder are trainable. To ensure stable training across arbitrary resolutions, we progressively scale the input resolution: starting from a fixed 256×256, then 512×512, and finally allowing flexible resolutions up to 512×512. For training efficiency, we employ a bucket-resolution strategy, batching samples with similar resolutions. 
The training corpus consists of 63M samples of various types with the data composition shown in Fig.~\ref{fig:data_composition}.

\noindent \textbf{Diffusion decoder training.} We incorporate an additional diffusion model to decode image for enhanced generation quality and efficient super-resolution. In this stage, the two encoders and codebooks from \tokenizer~are frozen, while the pixel decoder is replaced by a diffusion-based decoder to reconstruct and and upscales images by a factor of 2.
To support flexible resolutions, we predefine 11 commonly used aspect ratios: \{1:1, 3:4, 4:3, 2:3, 3:2, 1:2, 2:1, 1:3, 3:1, 1:4, 4:1\}. Each image is matched to the closest predefined aspect ratio and cropped accordingly.
Note that images that cropped more than 20\% of the original content is removed during training to preserve image integrity.
For high-resolution support and efficient training, we adopt a two-stage process: the first stage handles images with a total pixel count near $512^2$, while the second stage scales up to approximately $1024^2$. The training data consists of a 10M-image subset from our tokenizer dataset.

\noindent \textbf{MLLM training.}
Following ILLUME, the training procedure consists of three stages as below, where we progressively unfreeze more parameters and increase the complexity and variety of tasks.

\textit{Stage 1: Visual Embedding Initialization.} 
The primary goal of this stage is to initialize a good visual representation. 
We optimize vision-related components, namely the adapter, the vision embedding, and the LM head of the vision part, on image reconstruction and image captioning tasks. 
In this stage, we fix the image resolution as $256 \times 256$.

\textit{Stage 2: Unified Image-Text Alignment.} 
This stage focuses on image-text alignment to learn on multimodal data. 
We unfreeze the LLM and vision adaptor, with training data covering a variety of tasks, including text data, image caption data for both natural images and documents, text-to-image generation, and image editing data. 
This process contains two sub-stages: the first sub-stage fixed image resolution as $256 \times 256$ while the second one uses fixed image resolution as $512 \times 512$.

\textit{Stage 3: Supervised Fine-tuning.} 
After pretraining, we train the whole model with task-specific data to handle various multimodal understanding, generation, and editing tasks. In this stage, we leverage flexible resolution training. Specifically, for understanding tasks, it processes images with their naive resolutions, while for image generation and editing tasks, each input image is matched to a predefined aspect ratio and resize-and-crop accordingly with a total pixel count near $512^2$. Note that benefit to the diffusion decoder, our final model enables generation up to $1024 \times 1024$ resolution with multiple aspect ratios.

%% file: sec/4_experiments.tex
\section{Experiments}
\label{sec:experiments}

\subsection{Implementation Details} 
In our experiments, we use Qwen2.5~\cite{qwen2.5} as the base LLM. In our \tokenizer, the semantic encoder uses a pretrained QwenVIT~\cite{wang2024qwen2}. The semantic decoder consists of 4 attention blocks with 2D-RoPE. The pixel encoder and decoder follow MoVQGAN-based architecture~\cite{zheng2022movq} with the basic channel of 128 and 384, respectively. We use a codebook size of 32,768 for the semantic branch and 98,304 for the pixel branch. Codebook dimensions are 32 for both semantic and pixel codebooks. We apply AdamW optimizer~\cite{loshchilov2017adamw} without weight decay and constant learning rate for \tokenizer, diffusion decoder and MLLM.
The specific training hyperparameters of three parts are summarized in Table~\ref{tab:hyper_params}. 
The training process of \tokenizer~and the diffusion decoder took around 3+3 days on a cluster of 256 Ascend NPUs. Then 3B MLLM model took about 13 days to finish the 3-stage training.

\input{tables/experiments_hyper_params}

\subsection{Compare to State-of-the-Art}
\begin{figure*}[t]
\vspace{-5mm}
\centering
\includegraphics[width=0.95\textwidth]{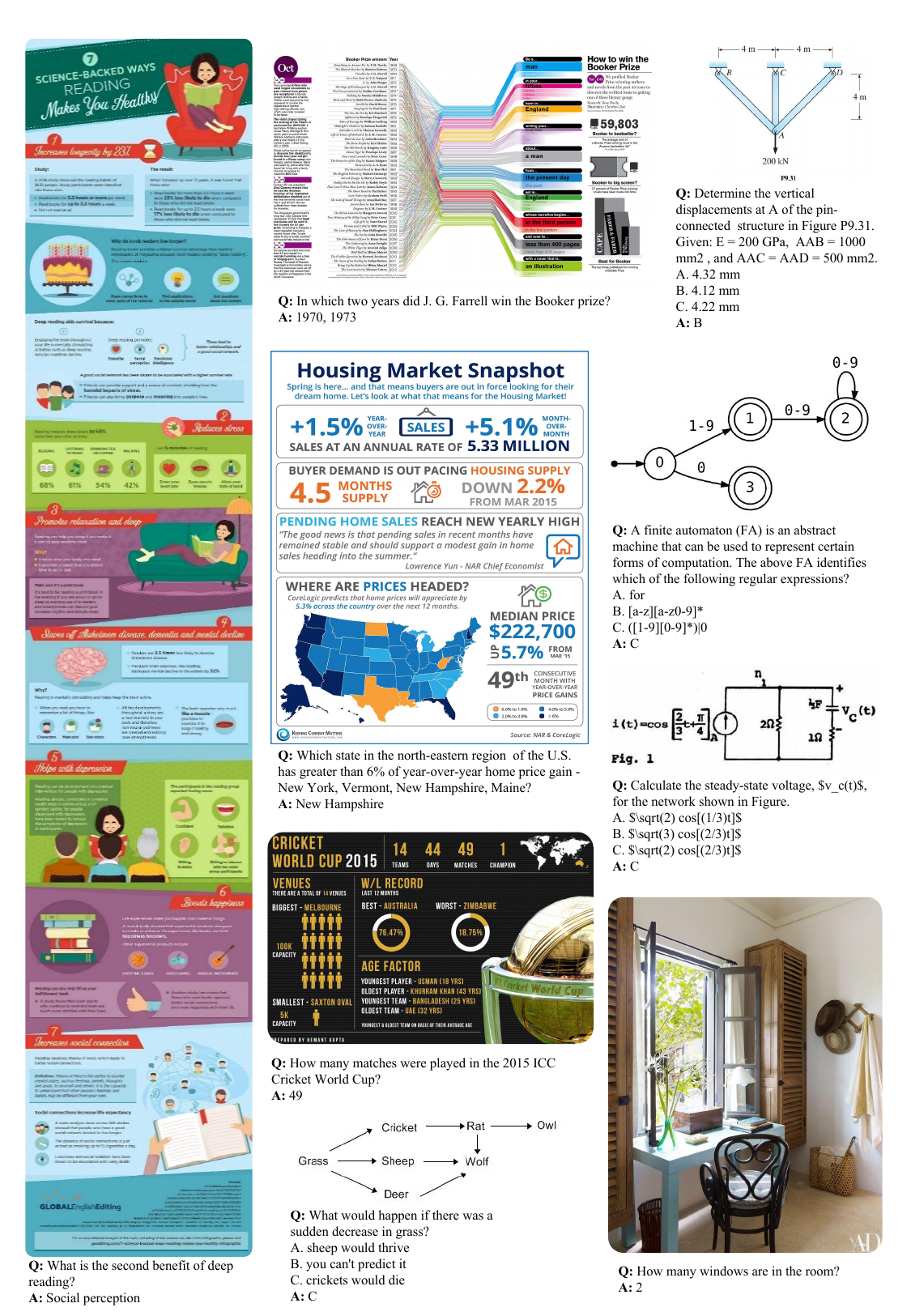}
\caption{More visualizations on understanding tasks.}
\vspace{-3mm}
\label{fig:vis_und}
\end{figure*}

\noindent \textbf{Multimodal understanding.}
To evaluate the multimodal understanding capabilities, we conduct evaluation on two types of widely-used benchmarks: (1) \textit{General}, including POPE~\cite{POPE}, MMBench~\cite{mmbench}, SEED~\cite{seed}, MME-P~\cite{mme}, MM-Vet~\cite{mmvet}, MMMU~\cite{mmmu} and AI2D~\cite{ai2d}; (2) \textit{Document-oriented}, including VQA-text~\cite{vqatext}, ChartQA~\cite{chartqa}, DocVQA~\cite{docvqa}, InfoVQA~\cite{infovqa} and OCRBench~\cite{ocrbench}. 
As shown in our results, despite using only a 3B model, ILLUME+ achieves competitive performance compared to state-of-the-art unified models, including Janus-Pro-7B and ILLUME-7B, on general benchmarks. 
Notably, our 3B model demonstrates exceptional performance on document-related tasks, a challenge for most existing unified models. This highlights the effectiveness of our dual-encoder design in preserving strong understanding capabilities within a unified model. More visualizations on understanding tasks are illustrated in Fig.~\ref{fig:vis_und}.

\input{tables/result_understanding}
\input{tables/result_generation}

\begin{figure*}[t]
\vspace{-5mm}
\centering
\includegraphics[width=0.96\textwidth]{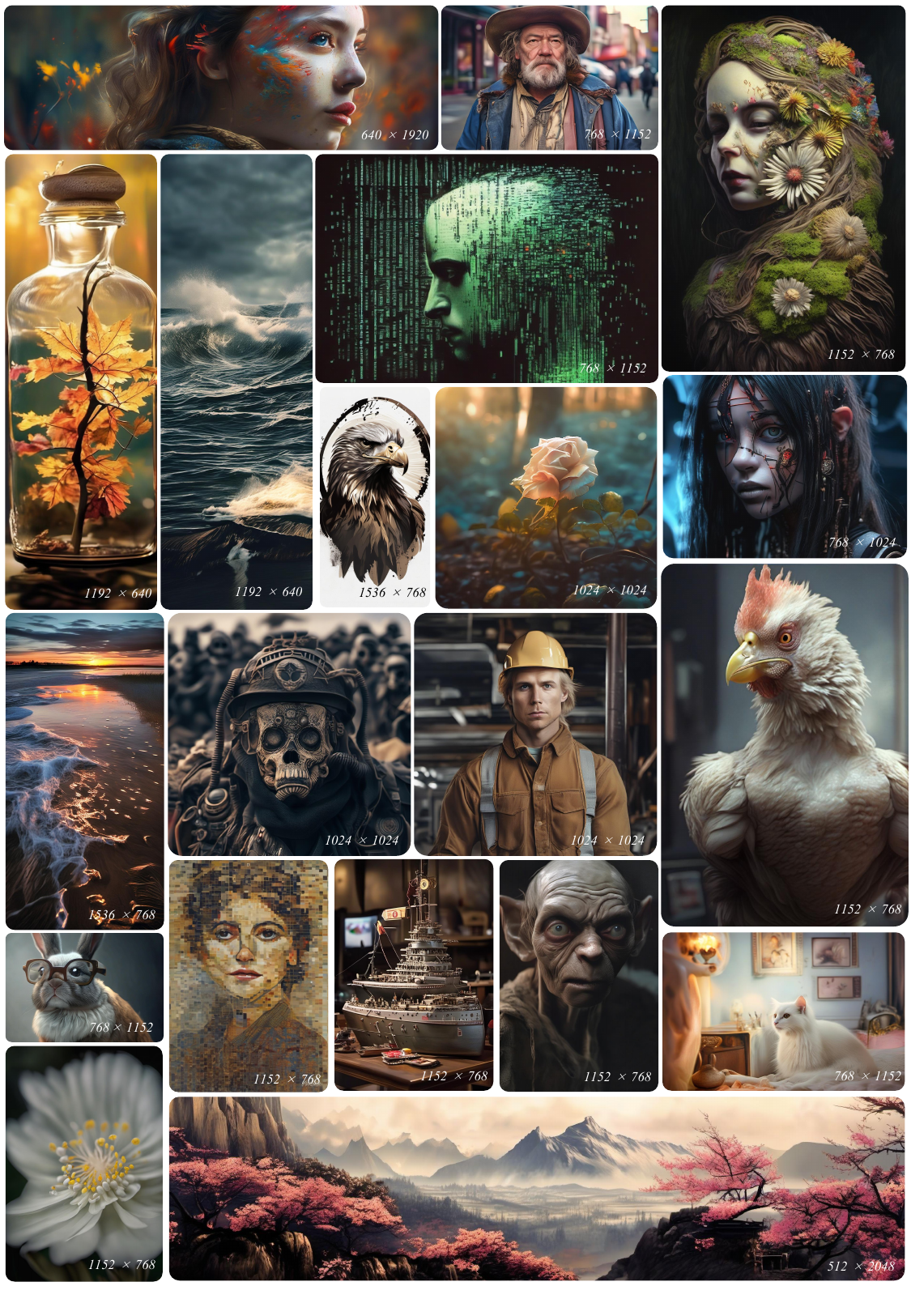}
\caption{More visualizations on generation tasks.}
\vspace{-3mm}
\label{fig:vis_gen}
\end{figure*}

\noindent \textbf{Multimodal image generation.}
To evaluate the multimodal visual generation capability, we use the MJHQ-30K~\cite{mjhq}, GenAI-bench~\cite{genaibench} and GenEval~\cite{geneval} benchmarks in Table~\ref{tab:generation_results}. For MJHQ-30K, we adopt the Fréchet Inception Distance (FID~\cite{fid}) metric on 30K generated images compared to 30K high-quality images, measuring the generation quality and diversity. GenAI-bench~\cite{genaibench} and GenEval~\cite{geneval} are challenging text-to-image generation benchmarks designed to reflect the consistency between text descriptions and generated images.
We compare \method~with previous state-of-the-art multimodal generation-only and unified models. With our dual vision tokenizer, ILLUME achieves a 6.00 FID score on the MJHQ30K benchmark, achieving state-of-the-art performance across both generation-only and unified models. This highlights the superior generation quality and diversity enabled by our diffusion-based approach. Additionally, \method~achieves competitive results on the GenAI-bench and GenEval benchmarks and attains the highest accuracy (0.72) in advanced categories on GenAI-bench, demonstrating its ability to understand and generate images from complex text descriptions. Figure~\ref{fig:vis_gen} shows more results of \method~on generating flexible resolution images.

\input{tables/result_tokenizer_compare}

\begin{figure*}[t]
\vspace{-5mm}
\centering
\includegraphics[width=0.97\textwidth]{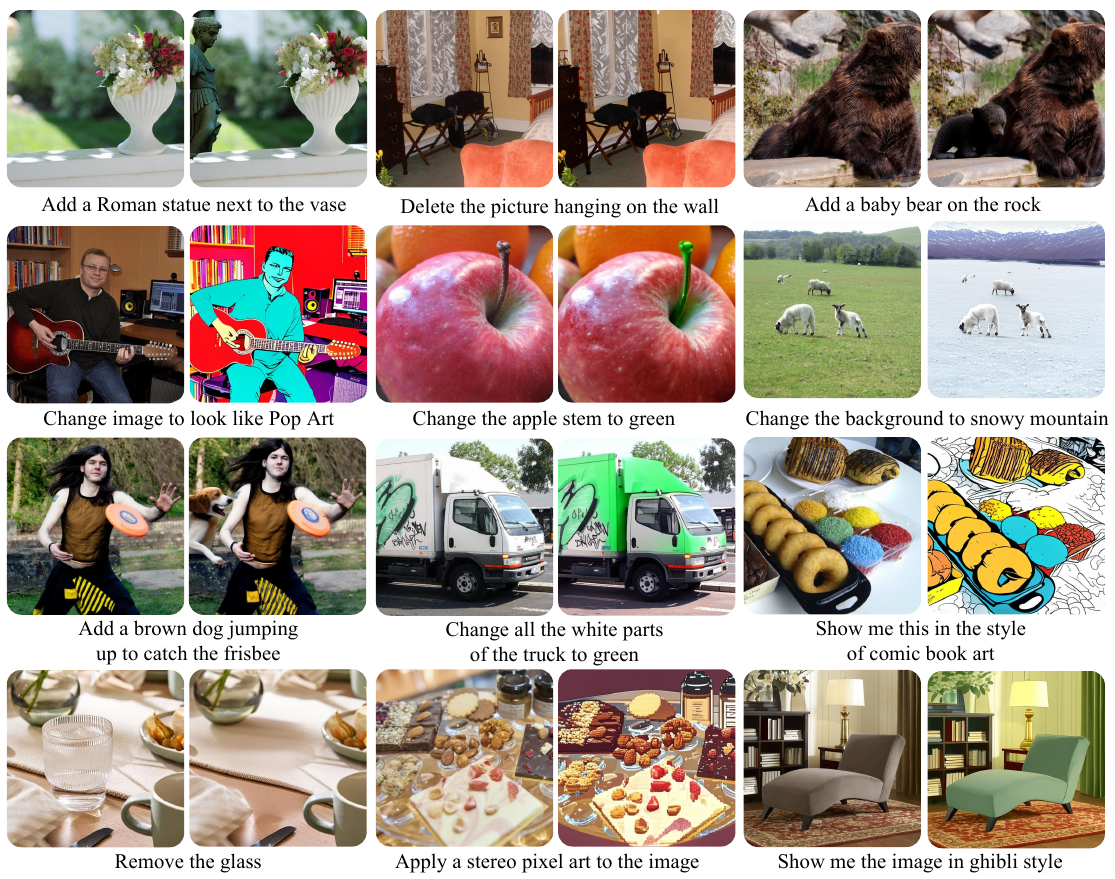}
\caption{More visualizations on editing tasks.}
\vspace{-3mm}
\label{fig:vis_edit}
\end{figure*}

\noindent \textbf{Multimodal image editing.}
To assess the multimodal image editing capability of our method, we evaluate it on the Emu Edit~\cite{emuedit} benchmark and report the CLIP-I, CLIP-T, CLIP-DIR and DINO~\cite{dino} scores. The CLIP-I and DINO scores measure the model’s ability to preserve elements from the source image, while the CLIP-T and CLIP-DIR score measures the consistency between the output image and the target caption. 
As illustrated in Table~\ref{tab:editing}, our model demonstrates strong performance in image editing tasks, surpassing specialized models, particularly in the CLIP-T metric. This indicates that the unified model's superior understanding enhances its ability to interpret editing instructions, resulting in more precise modifications. Furthermore, our dual-codebook design, which accounts for texture information, improves consistency with the original image as shown in Fig.~\ref{fig:vis_edit}.

\noindent \textbf{Image reconstruction of vision tokenizer.} Table~\ref{tab:tokenizer_compare} compares various state-of-the-art visual tokenizers on the ImageNet 50k validation set across different image resolutions using the rFID, PSNR, and SSIM.  At a resolution of $256\times256$, our \tokenizer~achieves state-of-the-art performance, exhibiting the best performance among the compared methods. Notably, \tokenizer~demonstrates the capability to handle multiple resolutions within a single model. For instance, when comparing performance at a higher resolution of 384x384, \tokenizer~significantly outperforms VILA-U~\cite{wu2024vila} at the same resolution, with a substantial improvement of 0.56 in rFID. This highlights \tokenizer~'s advantage in achieving superior reconstruction quality across varying input sizes with a single, versatile model, showcasing its efficiency and flexibility compared to fixed-resolution approaches like the specific VILA-U~\cite{wu2024vila} instance presented for 384x384.

\begin{table}[!t]
\centering
\tablestyle{2.5pt}{1}
\resizebox{1\textwidth}{!}{
\begin{tabular}{cccccccccccccc}
\toprule
    Continuous & Semantic & Pixel & \multicolumn{3}{c}{\textbf{Image Reconstruction}} & \textbf{Image Gen.} & \multicolumn{7}{c}{\textbf{Image Understanding}} \\ 
    Input & Encoder & Encoder & \textbf{rFID↓} & \textbf{PSNR↑} & \textbf{SSIM↑} & \textbf{gFID↓} & \textbf{POPE} & \textbf{MMB} & \textbf{SEED} & \textbf{MME-P} & \textbf{MM-Vet} & \textbf{MMMU} & \textbf{VQA-text} \\ 
\midrule
 \xmark & \cmark & \xmark & 5.48 & 15.69 & 0.487 &  -    & 81.6 & 46.9 & 54.1 & 1171.9 & 15.1 & 38.0 & 42.0  \\
 \xmark & \xmark & \cmark & 2.08 & 21.86 & 0.720 & 28.24 & 66.0 & 30.7 & 41.4 & 820.7 & 13.9 & 36.1 & 40.6 \\ 
 \xmark & \cmark & \cmark & 1.83 & 21.68 & 0.714 & 26.70 & 82.1 & 50.4 & 56.0
& 1203.7 & 17.7 & 39.7 & 43.3\\
\rowcolor{mygray} \cmark & \cmark & \cmark & 1.83 & 21.68 & 
0.714 & 26.70 & \textbf{85.3} & \textbf{70.9} & \textbf{66.6} & \textbf{1491.6} & \textbf{34.0} & \textbf{42.4} & \textbf{56.2}\\
\bottomrule   
\end{tabular}
}
\vspace{1mm}
\caption{\textbf{Ablation study of visual tokenizer on image reconstruction and image generation.} The rFID, PSNR and SSIM are evaluated on ImageNet 50k validation set. The gFID is evaluated on MJHQ30k.
The setting of the main experiments are marked in \mygrayhl{gray}.}
\label{tab:tokenizer_compare_semantic_pixel}
\end{table}

\begin{table}[t]
    \centering
    \begin{minipage}[t]{0.55\textwidth}
        \centering
        \vspace{0pt}
        \begin{minipage}[t]{\textwidth}
            \centering
            \tablestyle{2.5pt}{1}
            \resizebox{1\textwidth}{!}{
            \begin{tabular}{cccccccc}
                \toprule
                Encoder & Decoder & Up/down & Codebook &  &  &  &  \\ 
                Channel & Channel & Block & Size & Noise & \textbf{rFID↓} & \textbf{PSNR↑} & \textbf{SSIM↑} \\ 
                \midrule
                256 & 256 & Origin &  32k/32k & None & 1.83 & 21.68	& 0.714 \\
                128 & 256 & Origin &  32k/32k & None & 1.67 & 21.95 & 0.721 \\
                128 & 384 & Origin &  32k/32k & None & 1.54 & 21.98 & 0.723  \\
                128 & 384 & DC Block &  32k/32k & None & 1.41 & 21.76 & 0.716 \\
                128 & 384 & DC Block &  32k/98k & None & 1.44 & \textbf{22.29} & \textbf{0.736} \\
            \rowcolor{mygray} 128 & 384 & DC Block & 32k/98k & Random & \textbf{1.33} & 22.21 & 0.731\\
                \bottomrule   
            \end{tabular}
            }
            \vspace{1mm}
            \caption{\textbf{Ablation study of \tokenizer's components.} The first value of the Codebook Size is semantic codebook size and the second is the pixel codebook size. The setting of the main experiments are marked in \mygrayhl{gray}.}
            \label{tab:tokenizer_compare}
        \end{minipage}
        \vspace{1em} 
        \begin{minipage}[t]{\textwidth}
            \centering
            \tablestyle{2.5pt}{1}
            \resizebox{1\textwidth}{!}{
            \begin{tabular}{lccccc}
            \toprule
            Quantizer & Codebook Dim & \textbf{rFID↓} & \textbf{PSNR↑} & \textbf{SSIM↑} & Codebook Utilization \\ 
            \midrule
                VQ     & 32 & 2.24 & 20.81 & 0.666 & 6.59\%   \\
            \rowcolor{mygray} SimVQ  & 32 &  1.83 & 21.68 & 0.714 & 100\% \\
                SimVQ  & 16 & 1.82 & 21.89 & 0.721 & 100\% \\
                SimVQ  & 8  & 1.84 & 21.95 & 0.715&  100\%   \\
            \bottomrule   
            \end{tabular}
            }
            \vspace{1mm}
            \caption{\textbf{Ablation study of visual tokenizer about quantization types and codebook dimensions.} The setting of the main experiments are marked in \mygrayhl{gray}.}
            \label{tab:tokenizer_compare_codebook_dim}
        \end{minipage}
    \end{minipage}%
    \hfill
    \begin{minipage}[t]{0.42\textwidth}
        \centering
        \vspace{0pt}
        \tablestyle{2.5pt}{1}
        \resizebox{1\textwidth}{!}{
        \begin{tabular}{cccccc}
        \toprule
            Noise Type & $\alpha$ & $\beta$ & \textbf{rFID↓} & \textbf{PSNR↑} & \textbf{SSIM↑} \\ 
        \midrule
            -  & -  & -  & 1.83 & 21.68 & 0.714  \\
            Random  & 10 & 100  & 1.76 & 21.85 & 0.722 \\
            Random  & 10 & 50  & 1.87 &	21.60 & 0.713 \\
        \rowcolor{mygray} Random  & 10 & 10 & 1.81 & 21.98 & 0.722  \\
            Random  & 50 & 10  &  1.90 & 21.47 & 0.710  \\
            Random  & 100 & 10  & 2.06 & 21.65 & 0.707  \\
        
            Zero   & 10 & 100 & 1.88 & 21.57 & 0.717  \\
            Zero   & 10 & 50 & 1.73 & 21.97 & 0.725   \\
            Zero   & 10 & 10 & 1.90 & 21.97 & 0.722   \\
            Zero   & 50 & 10 & 1.75 & 21.88 & 0.722  \\
            Zero   & 100 & 10 & 1.79 & 21.75 & 0.716 \\
        \bottomrule   
        \end{tabular}}
        \vspace{1mm}
        \caption{\textbf{Ablation of noise on visual tokenizer.} The noise has $\alpha$\% probability to perturbe the current sample, with $\beta$\% of its tokens randomly replaced. The setting of the main experiments are marked in \mygrayhl{gray}. }
        \label{tab:tokenizer_noise_compare}
    \end{minipage}
    \vspace{-8mm}
\end{table}

\subsection{Ablation Studies}
In our ablation studies, we train our \tokenizer~on the ImageNet training set for 20 epochs at an image resolution of $256 \times 256$. Unless specified otherwise, both the semantic codebook and the pixel codebook have a vocabulary size of 32768, respectively. For image generation, we train the MLLM with \tokenizer~on a dataset of 10M high-fidelity images. For image understanding, we train the MLLM on LLaVA's pretraining and SFT datasets. The performance of image reconstruction is assessed on ImageNet val (50k), and the gFID is assessed on MJHQ30k.

\noindent \textbf{Dual tokenization vs. single tokenization.} We compare our \tokenizer~with the single tokenization method, i.e., semantic autoencoder and pixel autoencoder. Specifically, the semantic autoencoder is trained on both the image and semantic reconstruction task. The pixel autoencoder is only trained on image reconstruction. As shown in Table~\ref{tab:tokenizer_compare_semantic_pixel}, our dual tokenization, \tokenizer, which fuses the semantic and pixel branches, outperforms single tokenization methods in image reconstruction. Moreover, combining the semantic and pixel codebooks via the coarse-to-fine generation enhances the generation performance, i.e., a 1.54 rFID improvement. 
According to the image understanding abilities, the semantic encoder outperforms the pixel encoder, and our \tokenizer~boosts performance across all benchmarks.

\noindent \textbf{Continuous input vs. discrete input.}~The last two rows of Table~\ref{tab:tokenizer_compare_semantic_pixel} compare the performance differences between continuous visual input and discrete visual input. Obviously, continuous input leads to better performance on all benchmarks, proving the importance of continuous input in achieving superior image understanding ability.

\noindent \textbf{Ablation of the components of \tokenizer.} Table~\ref{tab:tokenizer_compare} demonstrates the components to improve our \tokenizer's baseline from 1.83 to 1.33 on rFID. First, we find the smaller encoder and larger decoder can improve the reconstruction performance from 1.83 to 1.54. Then the application of the DC block~\cite{chen2024dcae} brings 0.13 improvements. Scaling the pixel codebook size from 32k to 98k can further improve the performance on PSNR and SSIM while the import of random noise in the pixel codebook can improve the rFID to 1.33.

\noindent \textbf{Ablation of the random noise in \tokenizer.} Table~\ref{tab:tokenizer_noise_compare}  presents an ablation study on the impact of random noise and zero noise of the visual tokenizer as well as the effects of the $\alpha$ and $\beta$ as described in Sec.~\ref{sec:dual_vision_tokenizer}.
The results indicate that both random and zero noise can achieve better reconstruction performance compared to the baseline model. However, because random noise more accurately reflects the erroneous tokens predicted by LLMs it was chosen for the main experiments.

\noindent \textbf{Effect of the quantization method.} We compare vanilla VQ~\cite{esser2021taming} with SimVQ~\cite{zhu2024simvq} on Table~\ref{tab:tokenizer_compare_codebook_dim}. Results show that SimVQ can achieves better reconstruction performance on ImageNet and maintains high codebook utilization rate. Besides, we compare different codebook dimension and find similar performance between dimension of 32, 16 and 8.

%% file: tables/experiments_hyper_params.tex
\begin{table}[t]
\centering
\tablestyle{2.5pt}{1.2}
\resizebox{1\textwidth}{!}{
    
\begin{tabular}{l|c|c|cccc}
\toprule
 & \multirow{2}{*}{\textbf{\tokenizer}} & \multirow{2}{*}{\textbf{Diffusion Decoder}} & \multicolumn{4}{c}{\textbf{MLLM}} \\
\multirow{-2}[0]{*}{Settings} &  &  & Stage 1 & Stage 2-1 & Stage 2-2 & SFT \\ 
\midrule
    \multirow{2}{*}{Learning Rate} & \multirow{2}{*}{1e-4}  & \multirow{2}{*}{2e-5} & Vis.~Adapter 1e-3 & \multicolumn{2}{c}{Vis.~Adapter 5e-5} & Vis.~Adapter 2e-5   \\
     & & & Vis.~Embed.~\& Head 2e-4 & \multicolumn{2}{c}{LLM 5e-5} & LLM 2e-5, ViT 2e-6 \\
    Batch Size  & 256  & 128 & 1024 & 1024 & 512 & 256 \\
    Training Steps  & 270k/50k/78k & 265k & 5k & 98k & 40k & 40k \\
    Image Res. Mode  & Fix/Fix/Anyres & Multi-ratio & Fix & Fix & Fix & Anyres \\
    
    Image Main/Max Res.  & 256/512/512 & 512/1024 & 256 & 256 & 512 & 1024  \\
\bottomrule   
\end{tabular}
}
\vspace{1mm}
\caption{Training hyperpparameters of experiments.}
\vspace{-3mm}
\label{tab:hyper_params}
\end{table}

%% file: tables/result_understanding.tex
\begin{table*}[t]
\center
\tablestyle{3pt}{1.2}
\Large
\resizebox{\textwidth}{!}{
\begin{tabular}{ll|ccccccc|ccccc}
\toprule
 &   & \multicolumn{7}{c|}{\textbf{General}} & \multicolumn{5}{c}{\textbf{Doc}} \\
\multirow{-2}[0]{*}{Method} & \multirow{-2}[0]{*}{LLM.} & POPE & MMBench & SEED & MME-P & MM-Vet & MMMU & \multicolumn{1}{c|}{AI2D} & VQA-text & ChartQA & DocVQA & InfoVQA & OCRBench \\
\midrule
\multicolumn{14}{c}{\textbf{\it Understanding Only}} \\
\midrule
InstructBLIP~\cite{instructblip} & Vicuna-7B & - & 36.0 & 53.4 & - & 26.2  & 30.6  & 33.8 & 50.1 & 12.5  & 13.9  & - & 276 \\
Qwen-VL-Chat~\cite{bai2023qwen} & Qwen-7B & - & 60.6 & 58.2 & 1487.5 & -  & 35.9 & 45.9 & 61.5 &  66.3  & 62.6 & - & 488 \\
LLaVA-1.5~\cite{liu2024improved} & Vicuna-7B & 85.9 & 64.3 & 58.6 & 1510.7  & 31.1 & 35.4 & 54.8 & 58.2 & 18.2  & 28.1  & 25.8 & 318 \\
ShareGPT4V~\cite{chen2023sharegpt4v} & Vicuna-7B & - & 68.8 & 69.7 & 1567.4 & 37.6  & 37.2  & 58  & 60.4 & 21.3  & -  & - & 371 \\
LLaVA-NeXT~\cite{liu2024llava} & Vicuna-7B & 86.5 & 67.4 & 64.7 & 1519 & 43.9 & 35.1 & 66.6 & 64.9 & 54.8  & 74.4  & 37.1 & 532 \\
Emu3-Chat~\cite{wang2024emu3} & 8B from scratch & 85.2 & 58.5 & 68.2 & - & 37.2  & 31.6  & 70.0 & 64.7 & \underline{68.6} & \underline{76.3}  & 43.8 & \underline{687} \\
\midrule
\multicolumn{14}{c}{\textbf{\it Unify Understanding and Generation}} \\
\midrule
Unified-IO 2~\cite{lu2024unified} & 6.8B from scratch & \underline{87.7} & -  & 61.8 & - & - & - & - & - & -  & - & - & - \\
Chameleon~\cite{team2024chameleon} & 7B from scratch & - & - & - & - & 8.3 & 22.4 & -  & -  & - & - & - & - \\
LWM~\cite{liu2024world} & LLaMA-2-7B & 75.2 & - & - & - & 9.6 & - & - & 18.8 & -  & - & - & -   \\
Show-o~\cite{xie2024show} & Phi-1.5B & 73.8 & - & - & 948.4 & - & 25.1 & - & - & -  & - & - & -  \\
VILA-U (256)~\cite{wu2024vila} & LLaMA-2-7B & 83.9 & - & 56.3 & 1336.2 & 27.7 & - & -  & 48.3 & -  & - & - & -  \\
VILA-U (384)~\cite{wu2024vila} & LLaMA-2-7B & 85.8 & - & 59 & 1401.8 & 33.5 & - & -  & 60.8 & -  & - & - & -  \\
Janus~\cite{wu2024janus} & DeepSeek-LLM-1.3B & 87.0 & 69.4 & 63.7 & 1338.0 & 34.3 & 30.5 & -  & - & - & - & - & -  \\
Janus-Pro-1B~\cite{wu2024janus} & DeepSeek-LLM-1.3B & 86.2 & 75.5 & 68.3 & 1444.0 & 39.8 & 36.3  & -  & - & - & - & - & -  \\
Janus-Pro-7B~\cite{wu2024janus} & DeepSeek-LLM-7B &  87.4 & \underline{79.2} & \underline{72.1} & \textbf{1567.1} & \textbf{50.0} & \underline{41.0}  & -  & - & - & - & - & -  \\
\midrule
ILLUME & Vicuna-7B & \textbf{88.5} & 75.1 & 72.9 & \underline{1445.3} & 37.0 & 38.2 & \underline{71.4} & \textbf{72.1} & 66.7 & 76.0 & \textbf{45.5} & 669 \\
\rowcolor{backgroundcolor} \method & Qwen-2.5-3B & 87.6 & \textbf{80.8} & \textbf{73.3} & 1414.0 & \underline{40.3} & \textbf{44.3} & \textbf{74.2} & \underline{69.9} & \textbf{69.9} & \textbf{80.8} & \underline{44.1} &  \textbf{672} \\

\bottomrule
\end{tabular}
}
\caption{\textbf{Quantitative results on visual understanding benchmarks.} Our performance is close to and even outperforms both understanding only and unified models. The performance with top-1 and top-2 value are denoted in bold and underline respectively.}
\label{tab:understanding_results}
\end{table*}

%% file: tables/result_generation.tex
\begin{table}[t]
\center
\tablestyle{3pt}{1.1}
\LARGE
\resizebox{\linewidth}{!}{
\begin{tabular}{lcc|c|cc|ccccccc}
\toprule
 &   &  & \multicolumn{1}{c|}{\it MJHQ30k} & \multicolumn{2}{c|}{\it GenAI-bench} & \multicolumn{7}{c}{\it GenEval} \\
\multirow{-2}[0]{*}{Method} & \multirow{-2}[0]{*}{Params.} & \multirow{-2}[0]{*}{Type} & \multicolumn{1}{c|}{FID} & Basic & \multicolumn{1}{c|}{Advanced} & Overall & Single Obj & Two Obj. & Counting & Colors & Position & Color Attri. \\
\midrule
\multicolumn{13}{c}{\textbf{\it Generation Only}} \\
\midrule
SDv1.5 \cite{rombach2022high} & 0.9B & Diffusion & - & - & - & 0.43 & 0.97 & 0.38 & 0.35 & 0.76  & 0.04 & 0.06  \\
PixArt-$\alpha$ \cite{chen2023pixart} & 0.6B & Diffusion & \underline{6.14} & - & - & 0.48 & 0.98 & 0.5 & 0.44 & 0.8  & 0.08 & 0.07  \\
SDXL \cite{podell2023sdxl} & 2.6B & Diffusion & 9.55 & \textbf{0.83} & 0.63 & 0.55 & 0.98 & 0.74 & 0.39 & 0.85  & 0.15 & 0.23  \\
Emu3-Gen \cite{wang2024emu3} & 8B & Autoregressive & - & - & - & 0.54 & 0.98 & 0.71 & 0.34 & 0.81  & 0.17 & 0.21  \\
\midrule
\multicolumn{13}{c}{\textbf{\it Unify Understanding and Generation}} \\
\midrule
Chameleon \cite{team2024chameleon} & 7B & Autoregressive & - & - & - & 0.39 & - & - & - & -  & -  & -  \\
LWM \cite{liu2024world} & 7B & Autoregressive & 17.77 & 0.63 & 0.53 & 0.47 & 0.93 & 0.41 & 0.46 & 0.79  & 0.09  & 0.15  \\
Show-o \cite{xie2024show} &  1.5B & Autoregressive & 15.18 & 0.70 & 0.60 & 0.53 & 0.95 & 0.52 & 0.49 & 0.82  & 0.11  & 0.28  \\
VILA-U(256) \cite{wu2024vila} &  7B & Autoregressive & 12.81 & \underline{0.76} & \underline{0.64} & $-$ & $-$ & $-$ & $-$ & $-$  & $-$  & $-$  \\
VILA-U(384) \cite{wu2024vila} & 7B & Autoregressive & 7.69 & 0.73 & 0.61 & $-$ & $-$ & $-$ & $-$ & $-$  & $-$  & $-$  \\
Janus~\cite{wu2024janus} & 1.3B & Autoregressive & 10.1 & $-$ & $-$ & 0.61 & 0.97 & 0.68 & 0.3 & 0.84  & 0.46  & 0.42 \\
Janus-Pro-1B~\cite{chen2025janus_pro} & 1.3B & Autoregressive & - & $-$ & $-$ & \underline{0.73} & 0.98 & 0.82 & 0.51 & \underline{0.89} & \underline{0.65} & \underline{0.56} \\
Janus-Pro-7B~\cite{chen2025janus_pro} & 7B & Autoregressive & - & $-$ & $-$ & \textbf{0.80} & \textbf{0.99} & \textbf{0.89} & \underline{0.59} & \textbf{0.90} & \textbf{0.79} & \textbf{0.66} \\

\midrule
ILLUME~\cite{wang2024illume} & 7B & Autoregressive & 7.76 & 0.75 & 0.60 & 0.61 & \textbf{0.99} & 0.86 & 0.45 & 0.71 & 0.39 & 0.28  \\
\rowcolor{backgroundcolor} \method & 3B & Autoregressive & \textbf{6.00} & 0.72 & \textbf{0.71} & 0.72 & \textbf{0.99 }& \underline{0.88} & \textbf{0.62} & 0.84 & 0.42 & 0.53  \\
\bottomrule
\end{tabular}
}
\vspace{1mm}
\caption{\textbf{Quantitative results on text-to-image generation benchmarks.} \method~achieves comparable results with specialist models and unified MLLMs. The performance with top-1 and top-2 value are denoted in bold and underline respectively.
}
\label{tab:generation_results}
\end{table}

%% file: tables/result_tokenizer_compare.tex

\begin{table}[t]
    \centering
    \begin{minipage}{0.49\textwidth}
        \centering
        \LARGE
        \tablestyle{3pt}{1}
        \resizebox{1.0\textwidth}{!}{
        \begin{tabular}{lcccccccc}
            \toprule
            \textbf{Method} & Res. & ratio & \# Scales & \textbf{Dim} & \textbf{Size} & \textbf{rFID↓} & \textbf{PSNR↑} & \textbf{SSIM↑} \\ 
            \midrule
            VQGAN \cite{esser2021taming} & 256 & 16 & 1 & 256 & 16384 & 4.99 & 20.00 & 0.629 \\
            MaskGIT \cite{chang2022maskgit} & 256 & 16 & 1 & 256 & 1024 & 2.28 & $-$ & $-$ \\
            LLamaGEN \cite{sun2024autoregressive} & 256 & 16 & 1 & 32 & 16384 & 2.19 & 20.79 & 0.675 \\
            VILA-U \cite{wu2024vila} & 256 & 14.2 & 4 & 256 & 16384 & 1.80 & - & - \\
            VILA-U \cite{wu2024vila} & 384 & 14.2 & 16 & 256 & 16384 & 1.25 & - & - \\
            \midrule
            \rowcolor{backgroundcolor} \tokenizer & 256 & 16 & 2 & 32 & 131072 & 1.37 & 22.53 & 0.741 \\
            \rowcolor{backgroundcolor} \tokenizer & 384 & 16 & 2 & 32 & 131072 & 0.69 & 23.62 & 0.769  \\
            \rowcolor{backgroundcolor} \tokenizer & 512 & 16 & 2 & 32 & 131072 & 0.45 & 24.83 & 0.803  \\
            \bottomrule 
        \end{tabular}
        }
        \vspace{1mm}
        \caption{\textbf{Comparisons with other visual tokenizers.} The evaluations are on ImageNet 50k validation set under different image resolution.}
        \label{tab:tokenier_compare}
    \end{minipage}
    \hfill
    \begin{minipage}{0.49\textwidth}
        \centering
        \LARGE
        \tablestyle{3.2pt}{1}
        \resizebox{1.0\textwidth}{!}{
        \begin{tabular}{lcc|cccc}
            \toprule
             & & & \multicolumn{4}{c}{\it Emu Edit}  \\
            \multirow{-2}[0]{*}{Method} & \multirow{-2}[0]{*}{Type} & \multirow{-2}[0]{*}{Tasks} & DINO & CLIP-I & CLIP-T & CLIP-DIR \\
            \midrule
            InstructPix2Pix~\cite{brooks2023instructpix2pix} & Diffusion & Edit only & 0.762 & 0.834 & 0.219 & 0.078 \\
            MagicBrush~\cite{zhang2024magicbrush} & Diffusion & Edit only & 0.776 & 0.838 & 0.222 & 0.09\\
            OmniGen~\cite{xiao2024omnigen} & Diffusion & Edit only & 0.804 & 0.836 & 0.233 & -\\
            Emu Edit~\cite{emuedit} & Diffusion & Edit only & \underline{0.819} & 0.859 & 0.231 & \textbf{0.109}\\
            \midrule
            PUMA~\cite{fang2024puma} & AR & Edit only & 0.785 & 0.846 & \underline{0.270} & -\\
            ILLUME & AR & Und, Gen, Edit & 0.791 & \textbf{0.879} & 0.260 & - \\
            \rowcolor{backgroundcolor} \method~ & AR & Und, Gen, Edit & \textbf{0.826} & \underline{0.872} & \textbf{0.275} & \underline{0.101} \\
            \bottomrule
        \end{tabular}
        }
        \vspace{1mm}
        \caption{\textbf{Quantitative results on image editing benchmarks. } The performance with top-1 and top-2 value are denoted in bold and underline.
        }
        \label{tab:editing}
    \end{minipage}
\end{table}

%% file: sec/5_conclusion.tex
\section{Conclusion}
\label{sec:conclusion}
In this paper, we present \textbf{\method}, an enhanced version of the ILLUME model, which advances the integration of visual understanding, generation, and editing in a unified multimodal large language model. \method~proposes the dual visual tokenizor, \tokenizer, to preserve semantic and texture in images and utilizes a diffusion decoder to enhance image generation and achieve super-resolution. By leveraging unified coarse-to-fine image representation and a progressive training procedure for dynamic visual resolution, ILLUME+ with only 3B parameters enables to process flexible resolution visual inputs and outputs, and demonstrates good performance across various benchmarks in multimodal understanding, generation, and editing tasks.

There are several promising directions for future work, including scaling to larger model sizes (7B+) for enhanced task generalization and developing more advanced image-text interleaved pretraining techniques. Further improvements can be made by constructing more complex multimodal datasets and exploring post-training strategies for unified models. These advancements will help unlock the full potential of unified multimodal models in real-world applications, supporting more sophisticated tasks and enabling broader generalization across domains.